\documentclass[10pt,twocolumn,letterpaper]{article}

\usepackage{iccv}
\iccvfinalcopy 

\def\iccvPaperID{2179} 

\ificcvfinal\fi

\usepackage{times}
\usepackage{epsfig}
\usepackage{graphicx}
\usepackage{amsmath}
\usepackage{amssymb}

\usepackage[pagebackref=true,breaklinks=true,letterpaper=true,colorlinks,bookmarks=false]{hyperref}

\begin{document}

\title{Harvesting Discriminative Meta Objects with Deep CNN Features\\ for Scene Classification}

\author{$\mbox{Ruobing Wu\footnotemark[2]\hspace{1.5mm}}^1$\hspace{0.6in} Baoyuan Wang\footnotemark[3]\hspace{0.6in} Wenping Wang\footnotemark[2]\hspace{0.6in} Yizhou Yu\footnotemark[2]\\
\\
\footnotemark[2]\hspace{1mm} The University of Hong Kong \hspace{0.8in} \footnotemark[3]\hspace{1mm} Microsoft Technology and Research
}

\maketitle

\begin{abstract}
Recent work on scene classification still makes use of generic CNN features in a rudimentary manner. In this ICCV 2015 paper, we present a novel pipeline built upon deep CNN features to harvest discriminative visual objects and parts for scene classification. We first use a region proposal technique to generate a set of high-quality patches potentially containing objects, and apply a pre-trained CNN to extract generic deep features from these patches. Then we perform both unsupervised and weakly supervised learning to screen these patches and discover discriminative ones representing category-specific objects and parts. We further apply discriminative clustering enhanced with local CNN fine-tuning to aggregate similar objects and parts into groups, called meta objects. A scene image representation is constructed by pooling the feature response maps of all the learned meta objects at multiple spatial scales. We have confirmed that the scene image representation obtained using this new pipeline is capable of delivering state-of-the-art performance on two popular scene benchmark datasets, MIT Indoor 67~\cite{MITIndoor67} and Sun397~\cite{Sun397}.
\end{abstract}

\section{Introduction} \label{sec:intro}
\footnotetext[1]{This work was partially completed when the first author was an intern at Microsoft Research.}

Deep convolutional neural networks (CNNs) have gained tremendous attention recently due to their great success in boosting the performance of image classification \cite{DBLP:conf/nips/KrizhevskySH12,DBLP:conf/cvpr/OquabBLS14}, object detection \cite{RCNN:14,DBLP:journals/corr/SermanetEZMFL13}, action recognition \cite{DeepVideo14} and many other visual computing tasks \cite{DBLP:journals/corr/RazavianASC14,Pfister14a}.
In the context of scene classification, although a series of state-of-the-art results on popular benchmark datasets (MIT Indoor 67\cite{MITIndoor67}, SUN397 \cite{Sun397}) have been achieved, CNN features are still used in a rudimentary manner. For example, recent work in \cite{GigaSUN} simply trains the classical Alex's net \cite{DBLP:conf/nips/KrizhevskySH12} on a scene-centric dataset (``Places") and directly extracts holistic CNN features from entire images.

The architecture of CNNs suggests that they might not be best suited for classifying images, including scene images, where local features follow a complex distribution. The reason is that spatial aggregation performed by pooling layers in a CNN is too simple, and does not retain much information about local feature distributions. When critical inference happens in the fully connected layers near the top of the CNN, aggregated features fed into these layers are in fact global features that neglect local feature distributions.
It has been shown in \cite{DBLP:conf/eccv/GongWGL14} that in addition to the entire image, it is consistently better to extract CNN features from multiscale local patches arranged in regular grids.


In order to build a discriminative representation based on deep CNN features for scene image classification, we need to address two technical issues: (1) Objects within scene images could exhibit dramatically different appearances, shapes, and aspect ratios. To detect diverse local objects,
one could in theory add many perturbations to the input image by warping and cropping at various aspect ratios, locations, and scales, and then feed all of them to the CNN. This is, however, not feasible in practice; (2) To distinguish one scene category from another, it is much desired to harvest discriminative and representative category-specific objects and object parts. For example, to tell a ``city street" from a ``highway", one needs to identify objects that can only belong to a ``city street" but not a ``highway" scene. Pandey and Lazebnik~\cite{DBLP:conf/iccv/PandeyL11} adopt the standard DPM to adaptively infer potential object parts. It is however unclear how to initialize the parts and how to efficiently learn them using CNN features.

In this paper, we present a novel pipeline built upon deep CNN features for harvesting discriminative visual objects and parts for scene classification. We first use a region proposal technique to generate a set of high-quality patches potentially containing objects \cite{MCG:CVPR:14}. We apply a pre-trained CNN to extract generic deep features from these patches. Then, for each scene category, we train a one-class SVM on all the patches generated from the images for this class as a discriminative classifier~\cite{OCSVM:2001}, which heavily prunes outliers and other non-representative patches. The remaining patches correspond to the objects and parts that frequently occur in the images for this scene category. To further harvest the most discriminative patches, we apply a non-parametric weakly supervised learning model to screen these remaining patches according to their discriminative power across different scene categories. Instead of directly using the chosen category-specific objects and parts, we further perform discriminative clustering to aggregate similar objects and parts into groups. Each resulting group is called a \textbf{``Meta Object"}. Finally, a scene image representation is obtained by pooling the feature response maps of all the learned meta objects at multiple spatial scales to retain more information about their local spatial distribution. Locally aggregated CNN features are more discriminative than those global features fed into the fully connected layers in a single CNN.

There exists much recent work advocating the concept of middle-level objects and object parts for efficient scene image classification \cite{ObjectBank,DBLP:conf/iccv/PandeyL11,Singh:2012:UDM:2403006.2403013,juneja13blocks,DBLP:conf/nips/DoerschGE13,icml2013_wang13d}. Among them, the methods proposed in \cite{DBLP:conf/nips/DoerschGE13,juneja13blocks} are most relevant. Nonetheless, there exist major differences between our method and theirs. First, we use multiscale object proposals instead of grid-based sampling with multiple patch sizes, thus we can intrinsically obtain better discriminative object candidates. Second, we aggregate our meta objects through deep CNN features while previous methods primarily rely on low-level features (i.e., HOG). As demonstrated through experiments, deep features are more semantically meaningful when used for characterizing middle-level objects. Last but not the least, there exist significantly different components along individual pipelines. For instance, we adopt unsupervised learning to prune outliers while Juneja {\em et al.}~\cite{juneja13blocks} train a large number of exemplar-SVMs, which is more computationally intensive. Furthermore, our discriminative clustering component also plays an important role in aggregating meta objects.

In summary, this paper has the following contributions: (1) We propose a novel pipeline for scene classification that is built on top of deep CNN features. The advantages of this pipeline are orthogonal to any category independent region proposal methods \cite{SelectiveSearch:IJCV:13,EdgeBox:ECCV:14,MCG:CVPR:14} and middle-level parts learning algorithms \cite{DBLP:conf/nips/DoerschGE13,DBLP:conf/iccv/PandeyL11,juneja13blocks}. (2) We propose a simple yet efficient method that integrates unsupervised and weakly supervised learning for harvesting discriminative and representative category-specific patches, which we further aggregate into a compact set of groups, called meta objects, via discriminative clustering. (3) Instead of global fine-tuning, we locally fine-tune the CNN using the meta objects discovered from the target dataset. We have confirmed through experiments that the scene image representation obtained using this pipeline is capable of delivering state-of-the-art performance on two popular scene benchmark datasets, MIT Indoor 67~\cite{MITIndoor67} and Sun397~\cite{Sun397}.

\begin{figure*}[ht!]
  \centering
  \includegraphics[width=7in]{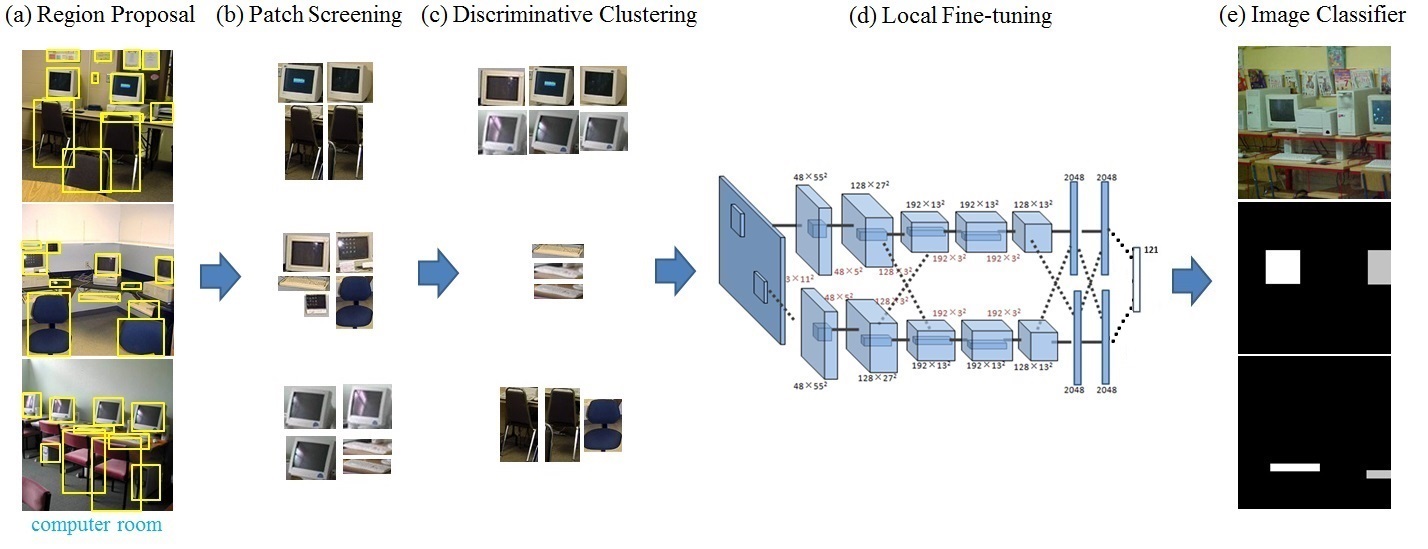}
  \caption{\label{im:flowgraph}Flowchart of our pipeline. From left to right: (a) Training scene images are processed by MCG \cite{MCG:CVPR:14} and we obtain top ranked region proposals (yellow boxes). (b) Patches are screened by our non-parametric scheme and only discriminative patches remain. (c) Discriminative clustering is performed to build meta objects. Three meta objects are shown here: `computer screen', `keyboard', `computer chair' (from top to bottom). Note that these names are for demonstration only, not labels applicable to our pipeline. (d) Local fine-tuning is performed on Hybrid CNN \cite{GigaSUN}, which decides which meta object a testing region belongs to. (e) We train an image classifier on aggregated responses of our fine-tuned CNN. Here the response maps of two meta objects, "computer screen" (second row) and "keyboard" (bottom row), are shown. Gray-scale values in the response maps indicate confidence. }
\end{figure*}

\section{A New Pipeline for Scene Classification} \label{Sec:Pipeline}
In this section, let us present the main components of our proposed new pipeline for scene classification. As illustrated in Figure \ref{im:flowgraph}, our pipeline is built on top of a pre-trained deep convolutional neural network, which is regarded as a generic feature extractor for image patches. In the context of scene classification, instead of directly transferring these features \cite{GigaSUN} or global fine-tuning on whole images using the groundtruth labels \cite{Decaf,RCNN:14}, we perform local fine-tuning on discriminative yet representative local patches that correspond to visual objects or their parts. As for scene classification datasets, bounding boxes or segment masks are not available for our desired local patches. In order to harvest them, we first adapt the latest algorithms to generate image regions potentially containing objects, expecting a high recall of all informative ones (Section \ref{Sec:Proposals}). Then we first apply an unsupervised learning technique, one-class SVMs, to prune those proposed regions that do not appear frequently in the images for a specific scene class. This is followed by a weakly supervised learning step to screen the remaining region proposals and discard those patches that are unlikely to be useful for differentiating a specific scene category from other categories (Section \ref{sec:MIL_Filter}).

To further improve the generality and representativeness of the remaining patches, we perform discriminative clustering to aggregate them into a set of meta objects (Section \ref{sec:cluster}). Finally, our scene image representation is built on top of the probability distribution of the mined meta objects (Section \ref{sec:final_representation}).

\subsection{Region Proposal Generation} \label{Sec:Proposals}
As discussed in Section \ref{sec:intro}, for arbitrary objects with varying size and aspect ratio, the traditional sliding window based object detection paradigm requires multiresolution scanning using windows with different aspect ratios. For example, in pedestrian detection \cite{Pedestrain:09}, at least two windows should be used to search for the full body and upper body of pedestrians. Recently, an alternative paradigm has been developed that performs perceptual grouping with the goal of proposing a limited number of high-quality regions, that likely enclose objects. Tasks including object detection \cite{RCNN:14} and recognition \cite{Recognition:Region} can then be built on top of these proposed regions only without considering other non-object regions. There is a large body of literature along this new paradigm for efficiently generating region proposals with a high recall, including selective search \cite{SelectiveSearch:IJCV:13}, edge-boxes \cite{EdgeBox:ECCV:14}, and multi-scale combinatorial grouping (MCG) \cite{MCG:CVPR:14}. We empirically choose MCG as the first component in our pipeline for generating high-quality region proposals, but one can use other methods as well. Figure \ref{im:objects} shows a few examples of regions generated by MCG. We also use region proposals from hierarchical image segmentation \cite{arbelaez2011contour} at the same time (see Sec.\ref{sec:Multi-level}).

\paragraph{Feature Extraction}
We use the CNN model pre-trained on the Places dataset \cite{GigaSUN} as our generic feature extractor for all the image regions generated by MCG. As this CNN model only takes input images with a fixed resolution, we follow the warping scheme described in R-CNN \cite{RCNN:14} and resample a patch with an arbitrary size and aspect ratio using the required resolution. Then each patch propagates through all the layers in the pre-trained CNN model, and we take the 4096-dimensional vector in the FC7 layer as the feature representation of the patch (see \cite{DBLP:conf/nips/KrizhevskySH12} and \cite{GigaSUN} for detailed information about the network architecture).

\subsection{Patch Screening} \label{sec:MIL_Filter}
\paragraph{Screening via One-Class SVMs} For each scene category, there typically exist a set of representative regions that frequently appear in the images for that category. For example, since regions with computer monitors frequently appear in the images for the ``computer room" class, a region containing monitors should be a representative region. Meanwhile, there are other regions that might only appear in few images. Such non-representative patches can be viewed as outliers for a certain scene category. On the basis of this observation, we adopt one-class SVMs \cite{OCSVM:2001} as discriminative models for removing non-representative patches. A one-class SVM separates all the data samples from the origin to achieve outlier detection. Let $x_1,x_2,...,x_l(x_i\in R^d)$ be the proposed regions from the same class, and $\Phi: X\longrightarrow H$ be a kernel function that maps original region features into another feature space. Training a one-class SVM needs to solve the following optimization:
\begin{eqnarray}\label{equ:ocsvm}
\min _{w,\xi,\rho} \frac{1}{2}\|w\|^2 + \frac{1}{\upsilon l}\sum_{i=1}^{l}\xi_i - \rho
\end{eqnarray}
subject to
\begin{eqnarray*}
(w\cdot\Phi(x_i)) \geq \rho - \xi_i,\; \xi_i \geq 0,\; i = 1,2,...,l,
\end{eqnarray*}
where $\upsilon (\in (0,1])$ controls the ratio of outliers. The decision function
\begin{equation}
f(x)=\mbox{sign}(w\cdot\Phi(x_i)-\rho)
\end{equation}
should return the positive sign given the representative patches and the negative sign given the outliers. This is because the representative patches tend to stay in a local region in the feature space while the outliers are scattered around in this space. To further improve the performance, we train a series of cascaded classifiers, each of which labels $15\%$ of the input patches as outliers and prune them. We typically use 3 cascaded classifiers.

\begin{figure}[t!]
  \centering
  \includegraphics[width=2.8in]{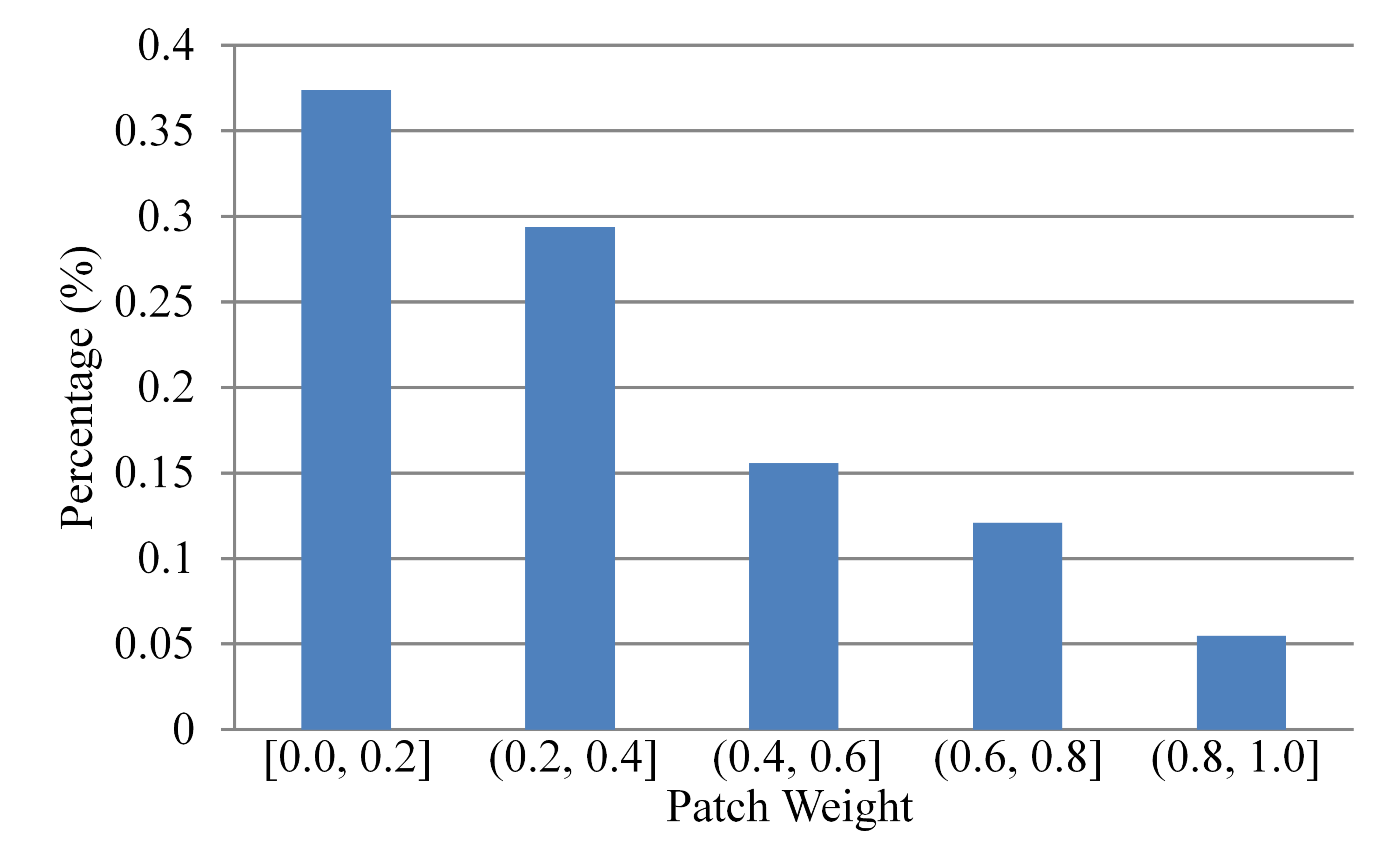}
  \caption{\label{im:weight}Patch weight distribution after weakly supervised patch screening. }
\end{figure}

\paragraph{Weakly Supervised Soft Screening}
After the region proposal step and outliers removal, let us suppose that $m_i$ image patches have been generated for each image $I_i$, and these patches likely contain objects or object parts. Let us denote a patch from $I_i$ as $p^i_j$ ($j \in \{1,...,m_i\}$), and use $y_i$ to represent the scene category label of image $I_i$. We associate each image patch $p^i_j$ with a weight $w_j^i \in [0,1]$ indicating the discriminative power of the patch among scene category labels. Our goal is to estimate this weight for every patch. Intuitively, a discriminative patch should have a high probability of appearing in one scene category and low probabilities of appearing in the other categories. That means, if we find the set of $K$ nearest neighbors $\textit{N}_{j}^i$ of $p^i_j$ from all image patches generated from all training images except $I_i$, we can use the following class density estimator to set $w_j^i$:
\begin{equation}\label{eq.MIL}
w_j^i = P(y_i|p^i_j) = \frac{P(p^i_j, y_i)}{P(p^i_j)} \approx K_y / K,
\end{equation}
where $K_y$ is the number of patches among the $K$ nearest neighbors that share the same scene label with $p^i_j$. By assuming that the $K$ nearest neighbors of $p^i_j$ are almost identical to $p^i_j$, we use $K_y$ to estimate the joint probability between a patch $p^i_j$ and its label $y_i$. Empirically we set $K$ to 100 in all the experiments. It is worth noting that patches with large weights also have more representative power. As representative patches would occur frequently in the visual world \cite{Singh:2012:UDM:2403006.2403013}, it is unlikely for non-representative patches to find similar ones (as its nearest neighbors) that share the same scene label. Fig. \ref{im:weight} shows the distribution of patch weights after our screening process.


\subsection{Meta Object Creation and Classification} \label{sec:cluster}
Once we have identified the most discriminative image patches, the next step is grouping these patches into clusters such that ideally patches in the same cluster should contain visual objects that belong to the same category and share the same semantic meaning. This is important for discovering the relationship between scene category labels and the labels of object clusters. Clustering also helps to show the internal variation of an object label. For example, desks facing a few different directions in a classroom might be grouped into several clusters. We call every patch cluster a meta object. Note that meta objects could correspond to visual objects but could also correspond to parts and patches that characterize the commonalities within a scene category.

\begin{figure*}[t!]
  \centering
    \includegraphics[width=0.64in, height=0.64in]{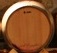}
    \includegraphics[width=0.64in, height=0.64in]{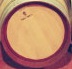}
    \includegraphics[width=0.64in, height=0.64in]{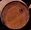}
    \includegraphics[width=0.64in, height=0.64in]{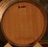}
    \includegraphics[width=0.64in, height=0.64in]{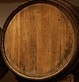}
    \includegraphics[width=0.64in, height=0.64in]{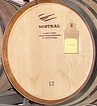}
    \includegraphics[width=0.64in, height=0.64in]{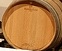}
    \includegraphics[width=0.64in, height=0.64in]{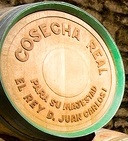}
    \includegraphics[width=0.64in, height=0.64in]{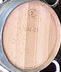}
    \hspace{0.4em}
    \includegraphics[width=0.64in, height=0.64in]{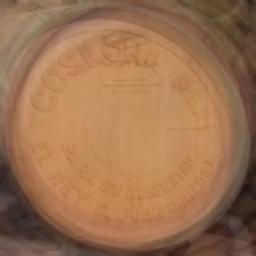}

    \includegraphics[width=0.64in, height=0.64in]{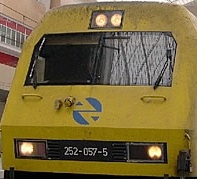}
    \includegraphics[width=0.64in, height=0.64in]{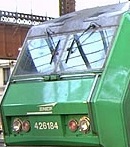}
    \includegraphics[width=0.64in, height=0.64in]{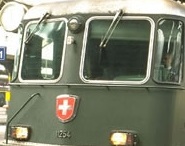}
    \includegraphics[width=0.64in, height=0.64in]{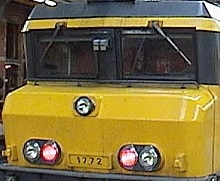}
    \includegraphics[width=0.64in, height=0.64in]{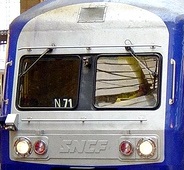}
    \includegraphics[width=0.64in, height=0.64in]{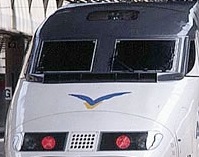}
    \includegraphics[width=0.64in, height=0.64in]{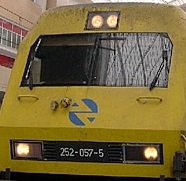}
    \includegraphics[width=0.64in, height=0.64in]{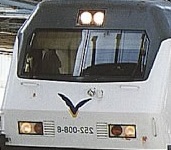}
    \includegraphics[width=0.64in, height=0.64in]{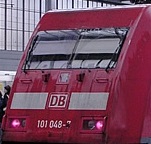}
    \hspace{0.4em}
    \includegraphics[width=0.64in, height=0.64in]{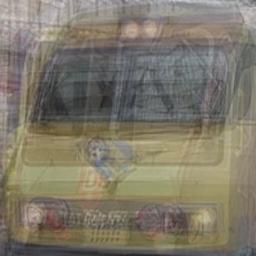}

    \includegraphics[width=0.64in, height=0.64in]{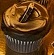}
    \includegraphics[width=0.64in, height=0.64in]{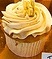}
    \includegraphics[width=0.64in, height=0.64in]{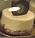}
    \includegraphics[width=0.64in, height=0.64in]{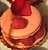}
    \includegraphics[width=0.64in, height=0.64in]{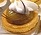}
    \includegraphics[width=0.64in, height=0.64in]{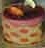}
    \includegraphics[width=0.64in, height=0.64in]{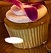}
    \includegraphics[width=0.64in, height=0.64in]{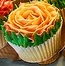}
    \includegraphics[width=0.64in, height=0.64in]{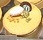}
    \hspace{0.4em}
    \includegraphics[width=0.64in, height=0.64in]{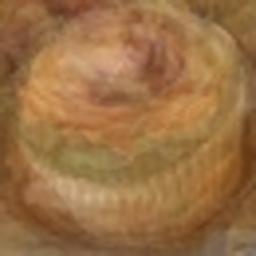}

    \includegraphics[width=0.64in, height=0.64in]{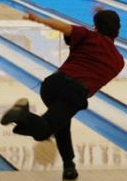}
    \includegraphics[width=0.64in, height=0.64in]{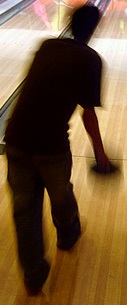}
    \includegraphics[width=0.64in, height=0.64in]{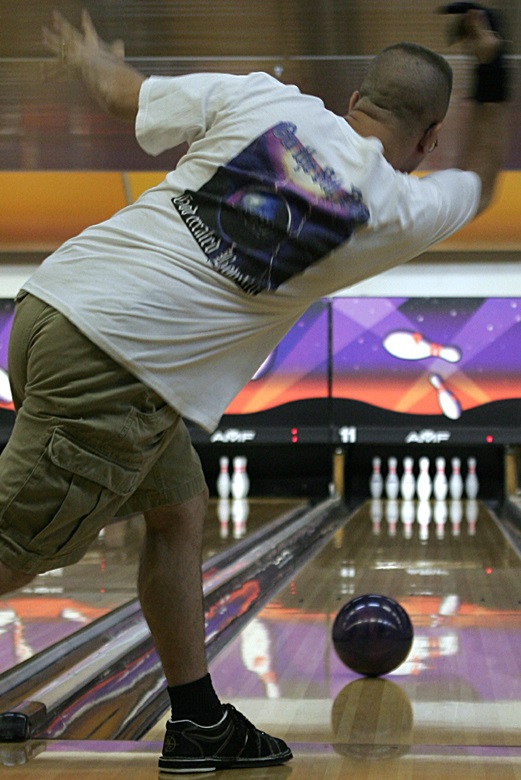}
    \includegraphics[width=0.64in, height=0.64in]{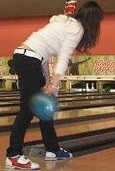}
    \includegraphics[width=0.64in, height=0.64in]{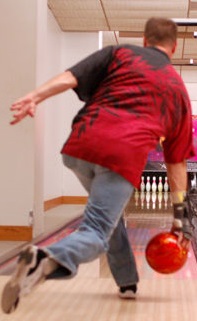}
    \includegraphics[width=0.64in, height=0.64in]{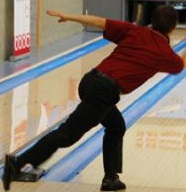}
    \includegraphics[width=0.64in, height=0.64in]{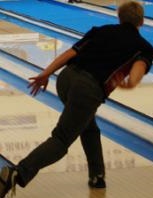}
    \includegraphics[width=0.64in, height=0.64in]{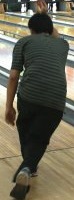}
    \includegraphics[width=0.64in, height=0.64in]{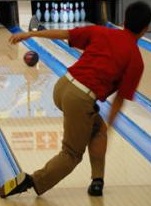}
    \hspace{0.4em}
    \includegraphics[width=0.64in, height=0.64in]{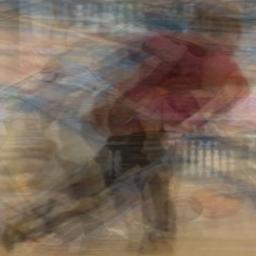}

    \includegraphics[width=0.64in, height=0.64in]{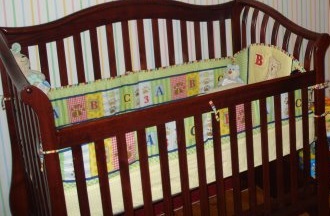}
    \includegraphics[width=0.64in, height=0.64in]{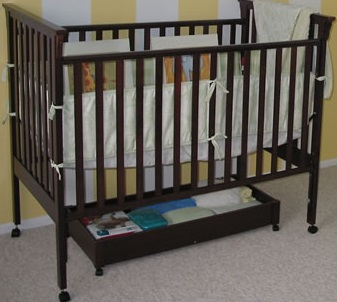}
    \includegraphics[width=0.64in, height=0.64in]{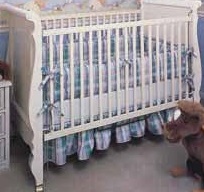}
    \includegraphics[width=0.64in, height=0.64in]{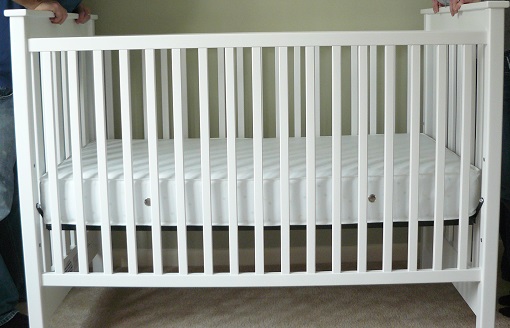}
    \includegraphics[width=0.64in, height=0.64in]{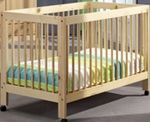}
    \includegraphics[width=0.64in, height=0.64in]{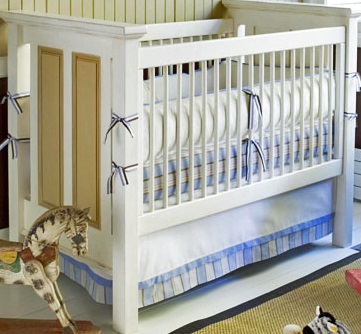}
    \includegraphics[width=0.64in, height=0.64in]{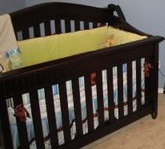}
    \includegraphics[width=0.64in, height=0.64in]{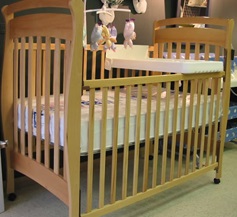}
    \includegraphics[width=0.64in, height=0.64in]{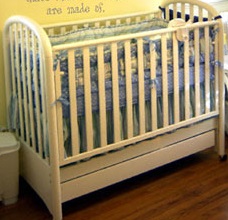}
    \hspace{0.4em}
    \includegraphics[width=0.64in, height=0.64in]{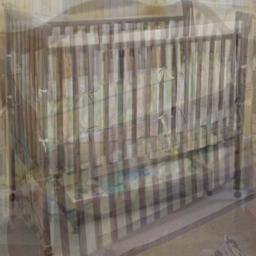}

    \includegraphics[width=0.64in, height=0.64in]{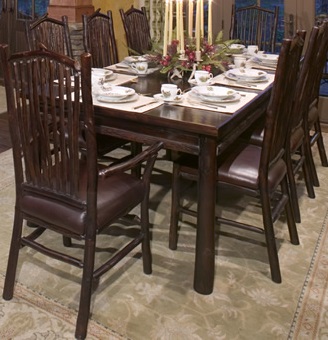}
    \includegraphics[width=0.64in, height=0.64in]{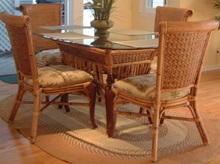}
    \includegraphics[width=0.64in, height=0.64in]{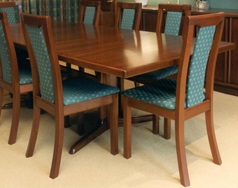}
    \includegraphics[width=0.64in, height=0.64in]{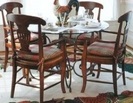}
    \includegraphics[width=0.64in, height=0.64in]{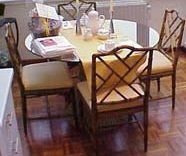}
    \includegraphics[width=0.64in, height=0.64in]{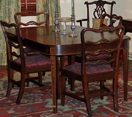}
    \includegraphics[width=0.64in, height=0.64in]{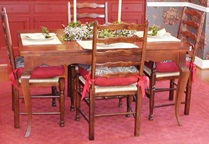}
    \includegraphics[width=0.64in, height=0.64in]{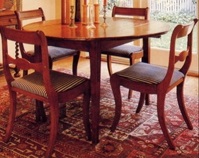}
    \includegraphics[width=0.64in, height=0.64in]{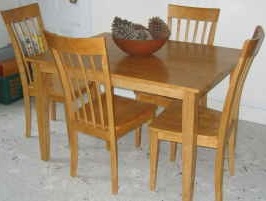}
    \hspace{0.4em}
    \includegraphics[width=0.64in, height=0.64in]{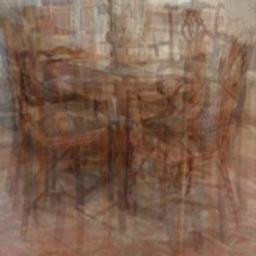}
  \caption{\label{im:objects}Examples of patch clusters (meta objects) from the MIT 67 Indoor dataset~\cite{MITIndoor67}. Patches on the same row belong to the same meta object. The rightmost column shows the average image, namely the `center', of the corresponding meta object.}
\end{figure*}

We adopt the Regularized Information Maximization (RIM) algorithm~\cite{Krause2010} to perform discriminative clustering. RIM strikes a balance among cluster separation, cluster balance and cluster complexity. Fig. \ref{im:objects} shows a few clusters after applying RIM to the screened discriminative patches from the MIT 67 Indoor Scenes dataset~\cite{MITIndoor67}. As we can see, the patches within the same cluster has similar appearances and the same semantic meaning. Here we can also observe the discriminative power of such clusters. For example, the wine buckets (top row in Fig. \ref{im:objects}) only show up in wine cellars, and the cribs (second row from the bottom in Fig. \ref{im:objects}) only show up in nurseries.

\paragraph{Local Fine-Tuning for Patch Classification} \label{Sec:Fine-tune}
Given the set of meta objects, we need a classifier to decide which meta object a patch from a testing image belongs to. There are various options for this classifier, including GMM-type probabilistic models, SVMs, and neural networks. We choose to fine-tune the pre-trained CNN on our meta objects, which include the collection of discriminative patches surviving the patch screening process.
We perform stochastic gradient descent over the pre-trained CNN using the warped discriminative image patches and their corresponding meta object labels. Take MIT Indoor 67 \cite{MITIndoor67} as an example. After weakly supervised patch screening (Section \ref{sec:MIL_Filter}), there exist around a million remaining image patches, and 120 meta objects are discovered during the clustering step (Section \ref{sec:cluster}). In the CNN, we replace the original output layer that performs ImageNet-specific 1000-way classification with a new output layer that does 121-way classification while leaving all other layers unchanged. Note that we need to add one extra class to represent those patches that are discarded during the screening step. The reason for local fine-tuning is obtaining an accurate meta object classifier that is also robust to noisy labels generated by the discriminative clustering algorithm used in Section \ref{sec:cluster}.

\subsection{Image Representation with Meta Objects} \label{sec:final_representation}
Inspired by previous work such as object-bank~\cite{ObjectBank} and bag-of-parts (BOF)~\cite{juneja13blocks}, we hypothesize that any scene image can be represented as a bag of meta objects as well.
Suppose $N$ meta objects have been learned during discriminative clustering (Section \ref{sec:cluster}).

Given a testing image, we still perform MCG to obtain region proposals. Every region can be classified into one of the discriminative object clusters using our meta object classifier. Spatial aggregation of these meta objects can be performed using Spatial Pyramid Matching (SPM)~\cite{yang2009linear}. In our implementation, we use three levels of SPM, and adaptively choose the centroid of all meta objects falling into a SPM region as the splitting center of its subregions. This strategy can better balance the number of meta objects that fall into each subregion. After applying SPM to the testing image, we obtain a hierarchical spatial histogram of meta object labels over the image, which can be used for determining the scene category of this testing image.

Another pooling method we consider is Vector of Locally Aggregated Descriptors (VLAD)~\cite{jegou2010aggregating, arandjelovic2013all}. We compute a modified version of VLAD that suits our framework. Specifically, we use our discriminative object clusters (meta objects) as the clusters for computing VLAD. That means we do not perform K-means clustering for VLAD. It is important to maintain such consistency because otherwise the recognition performance would degrade by 1.5\% on the MIT 67 Indoor Scenes dataset (from 78.41\% to 76.9\%). Other steps are similar to the standard VLAD. Given region proposals of an image, we assign each region to its nearest cluster center, and aggregate the residuals of the region features, resulting in a 4096-d vector per cluster. Suppose there are $k$ clusters. The dimension of this per-cluster vector is reduced to (4096/$k$)-d using PCA. Finally, these (4096/$k$)-d vectors are concatenated into a 4096-d VLAD descriptor.

The holistic Places CNN feature extracted from the whole image is also useful for training the scene image classifier since they also encode local as well as global information of the scene.

We train a neural network with two fully-connected hidden layers (each with 200 nodes) using normalized VLAD (or SPM) features concatenated with the holistic Places CNN features. The relative weight between these two types of features are learned via cross validation on a small portion of the training data. We use the rectified linear function (ReLU) as the activation function of the neurons in the hidden layers.

\subsection{Multi-Level Image Representation} \label{sec:Multi-level}
Our image representation with meta objects can be generalized to a multi-level representation. The insight here is that objects with different sizes and scales may supply complementary cues for scene classification. To achieve this, we switch to multi-level region proposals. The coarser levels deal with larger objects, while the finer levels deal with smaller objects and object parts. On each level, region proposals are generated and screened separately. Local fine-tuning for patch classification is also performed on each level separately. During the training stage of the final image classifier, the image representation is defined as the concatenation of the feature vectors from all levels. In practice, we find a 2-level representation sufficient. The bottom level includes relatively small regions from a finer level in a region hierarchy~\cite{arbelaez2011contour} to capture small objects and object parts in an image, while the top level includes region proposals generated by MCG as well as relatively large regions from a coarser level in the region hierarchy to capture large objects.


\section{Experiments and Discussions}
In this section, we evaluate the performance of our framework, named MetaObject-CNN, on the MIT Indoor 67 \cite{MITIndoor67} and SUN 397 \cite{Sun397} datasets as well as analyze the effectiveness of the specific choices we made at every stage of our pipeline introduced in Section \ref{Sec:Pipeline}.

\subsection{Datasets}
\paragraph{MIT Indoor 67} MIT Indoor 67 \cite{MITIndoor67} is a challenging indoor scene dataset, which contains 67 scene categories and a total of 15,620 images. The number of images varies across categories (but at least 100 images per category). Indoor scenes tend to have more variations in terms of composition, and are better characterized by the objects they have. This is consistent with the motivation of our framework.

\paragraph{SUN397} SUN397 \cite{Sun397} is a large-scale scene dataset, which contains 397 scene categories and a total of 108,754 images (also at least 100 images per category). The categories include different kinds of indoor and outdoor scenes which show tremendous object and alignment variance, thus bringing more complexity in learning a good classifier.

\subsection{Experimental Setup} \label{sec:setup}
For MIT Indoor 67, we train our model on the commonly adopted benchmark, which contains 80 training images and 20 testing images per category. There are 192 top ranked region proposals generated with MCG and 32 (96) regions from hierarchical image segmentation in the top (bottom) level for every training and testing image. The feature representation of a proposed region is set to the 4096-dimensional vector at the FC7 layer of the Hybrid CNN from \cite{GigaSUN}. After outlier removal (3 iterations of 15\% filtering out), we further discard 16\% patches, where the ratio is determined via cross validation on a small portion of the training data. Then we perform data augmentation (to 4 times larger) on the remaining patches using reflection, small rotation and random distortion. Discriminative clustering is performed on the augmented patches to produce 120 (40) meta objects for local fine-tuning in the bottom (top) level, which is performed on the Hybrid CNN by replacing the original output layer that performs ImageNet-specific 1000-way classification with a new output layer that does 121-way (41-way) classification while leaving all other layers unchanged. The pooling step (SPM and our modified VLAD) is discussed in Section \ref{sec:final_representation}. The image classification is done by a neural network with two fully-connected layers (200 nodes each) on the concatenated feature vector of VLAD pooling and the Hybrid CNN feature of the whole image.

For SUN 397, we adopt the commonly used evaluation benchmark that contains 50 training images and 50 testing images per category for each split from \cite{Sun397}. There are 96 top ranked regions generated with MCG and 32 (96) regions from hierarchical image segmentation in the top (bottom) level for every training and testing image. The feature representation of a proposed region is also set to the 4096-dimensional vector at the FC7 layer of the Hybrid CNN. After outlier removal (3 iterations of 15\% filtering out), we further discard 24\% patches. Data augmentation is also performed on the remaining patches involving reflection, small rotation and random distortion. Discriminative clustering results in 450 (150) meta objects in the bottom (top) level. Local fine-tuning is further performed on the Hybrid CNN by replacing the original output layer with a new output layer that does 451-way (151-way) classification while leaving all other layers unchanged. We also train a neural network with two fully-connected layers (200 nodes each) on the concatenated feature vector of VLAD pooling and the Hybrid CNN feature of the whole image to deal with image level classification.

\subsection{Comparisons with State-of-the-Art Methods}
In Table \ref{tab:MIT67}, we compare the recognition rate of our method (MetaObject-CNN) against published results achieved with existing state-of-the-art methods on MIT Indoor 67. Among the existing methods, oriented texture curves (OTC) \cite{margolin2014otc}, spatial pyramid matching (SPM) \cite{Lazebnik2006_SPM}, and Fisher vector (FV) with bag of parts \cite{juneja13blocks} represent effective feature descriptors as well as their associated pooling schemes. Discriminative patches \cite{Singh:2012:UDM:2403006.2403013,DBLP:conf/nips/DoerschGE13} are focused on mid-level features and representations. More recently, deep learning and deep features have proven to be valuable to scene classification as well~\cite{DBLP:conf/eccv/GongWGL14,GigaSUN}. The recognition accuracy of our method outperforms the state of the art by around 8.1\%.

\begin{table}[h]
\centering
\caption {Scene Classification Performance on MIT Indoor 67} \label{tab:MIT67}
\begin{tabular}{lc}
\\\hline
Method                                                           & Accuracy(\%)   \\ \hline
SPM \cite{Lazebnik2006_SPM}                                      & 34.40          \\
OTC \cite{margolin2014otc}                                       & 47.33          \\
Discriminative Patches ++ \cite{Singh:2012:UDM:2403006.2403013}  & 49.40          \\
FV + Bag of parts \cite{juneja13blocks}                          & 63.18          \\
Mid-level Elements \cite{DBLP:conf/nips/DoerschGE13}             & 66.87          \\
MOP-CNN \cite{DBLP:conf/eccv/GongWGL14}                          & 68.88          \\
Places-CNN \cite{GigaSUN}                                        & 68.24          \\
Hybrid-CNN \cite{GigaSUN}                                        & 70.80          \\
\textbf{MetaObject-CNN} & \textbf{78.90}      \\ \hline
\end{tabular}
\end{table}

Table. \ref{tab:SUN397} shows a comparison between the recognition rate achieved with our method (MetaObject-CNN) and those achieved with existing state-of-the-art methods on the SUN397 dataset. In addition to the methods introduced earlier, there exists additional representative work here. Xiao {\em et al.}~\cite{GigaSUN}, as the collector of SUN397, integrated 14 types of distance kernels including bag of features and GIST. DeCAF \cite{Decaf} uses the global 4096D feature from a pre-trained CNN model on ImageNet. OTC together with the HOG2x2 descriptor \cite{margolin2014otc} outperforms dense Fisher vectors~\cite{sanchez2013image}, both of which are effective feature descriptors for SUN397. And again, by applying deep learning techniques, MOP-CNN~\cite{DBLP:conf/eccv/GongWGL14} and Places-CNN~\cite{GigaSUN} (fine-tuned on SUN397) achieve state-of-the-art results (51.98\% and 56.2\%). With our MetaObject-CNN pipeline, we manage to achieve a higher recognition accuracy.

\begin{table}[h]
\centering
\caption {Scene Classification Performance on SUN397} \label{tab:SUN397}
\begin{tabular}{lc}
\\ \hline
Method                                                           & Accuracy(\%)   \\ \hline
OTC \cite{margolin2014otc}                                       & 34.56          \\
Xiao et al. \cite{GigaSUN}                                       & 38.00          \\
DeCAF \cite{Decaf}                                               & 40.94          \\
FV \cite{sanchez2013image}                                       & 47.20          \\
OTC+HOG2x2 \cite{margolin2014otc}                                & 49.60          \\
MOP-CNN  \cite{DBLP:conf/eccv/GongWGL14}                         & 51.98          \\
Hybrid-CNN \cite{GigaSUN}                                        & 53.86          \\
Places-CNN \cite{GigaSUN}                                        & 56.20          \\
\textbf{MetaObject-CNN} & \textbf{58.11}      \\ \hline
\end{tabular}
\end{table}

\subsection{Evaluation and Discussion}
In this section, we perform an ablation study to analyze the effectiveness of individual components in our pipeline. When validating each single component, we keep all the others fixed. Specifically, we treat the final result from our MetaObject-CNN as the baseline, and perform the analysis by altering only one component at a time. Table \ref{tab:inner} shows a summary of the comparison results on MIT Indoor 67. A detailed explanation of these results is given in the rest of this section.

\begin{table}[h]
\centering
\caption {Evaluation results on MIT Indoor 67 for varying pipeline configurations.} \label{tab:inner}
\begin{tabular}{lc}
\\ \hline
Configuration                                                               & Accuracy(\%)   \\ \hline \hline
Global fine-tuning                                  & 73.88          \\
\hline
Mode-seeking \cite{DBLP:conf/nips/DoerschGE13} with Hybrid-CNN     & 69.70          \\
Mode-seeking elements instead of MCG                                        & 76.34          \\
Dense grid-based patches                                                & 71.43          \\
\hline
Without outlier removal and patch screening                                 & 75.12          \\
Without outlier removal                                                     & 76.30          \\
Without patch screening                                                     & 78.82          \\
\hline
Without clustering                                                          & 72.81          \\
\hline
Without local fine-tuning                                                   & 76.10          \\
\hline
Cross-dataset evaluation                                                   & 76.52          \\
\hline
\textbf{MetaObject-CNN} & \textbf{78.90}      \\ \hline
\end{tabular}
\end{table}

\paragraph{Global vs Local Fine-Tuning}
Most of the previous methods \cite{GigaSUN,Decaf,DeepVideo14} using a pre-trained deep network primarily focus on global fine-tuning for domain adaptation tasks, which take the entire image as input and rely on the network itself to learn all the informative structures embedded within a new dataset. However, in this work, we perform fine-tuning on local meta objects harvested in an explicit manner. To compare, we start with the Places CNN network \cite{GigaSUN}, and fine-tune this network on MIT Indoor 67.
The recognition rate after such global fine-tuning is 73.88\% (top row in Table. \ref{tab:inner}), which is around 5\% lower than that of our pipeline. This indicates the advantages of our local approach of harvesting meta objects and performing recognition on top of them.
\vspace{-3mm}
\paragraph{Choice of Region Proposal Method} In addition to choosing MCG \cite{MCG:CVPR:14} and hierarchical image segmentation for generating object proposals, one might directly use dense grid-based patches or mid-level discriminative patches discovered by the pioneering techniques in \cite{DBLP:conf/nips/DoerschGE13,Singh:2012:UDM:2403006.2403013} as local object proposals. To evaluate the effectiveness of MCG, we have conducted the following three internal comparisons.


First, we compare our patch screening on top of region proposal with the patch discovery process in \cite{DBLP:conf/nips/DoerschGE13}, which is a piece of representative work on learning mid-level patches in a supervised manner. For a fair comparison, we use the Places CNN feature (FC7) to represent the visual elements in this work. Similar to the configuration in \cite{DBLP:conf/nips/DoerschGE13}, 1600 elements are learned per class and 200 top elements per class are used for further classification. The resulting recognition rate is 69.70\% (second row in Table. \ref{tab:inner}, which is 9.2\% lower than our result. This comparison demonstrates that region proposal plus patch screening is helpful in finding visual objects that characterize scenes. In a second experiment, we feed the top visual elements identified by \cite{DBLP:conf/nips/DoerschGE13} to our patch clustering step, and obtain 96 meta objects. The final recognition rate achieved with these meta objects is 76.34\% (third row in Table. \ref{tab:inner}), which is around 2.6\% lower than our result. This second experiment shows that MCG works with our pipeline better than mode-seeking elements from \cite{DBLP:conf/nips/DoerschGE13}. Then in a third experiment, instead of taking region proposals, we have tried using all patches from a regular 8x8 grid, the result is 71.43\% (fourth row in Table. \ref{tab:inner}), which indicates patches sampled from a regular grid are not good candidates for meta objects.

\begin{figure}[ht!]
  \centerline{\includegraphics[width=3.0in]{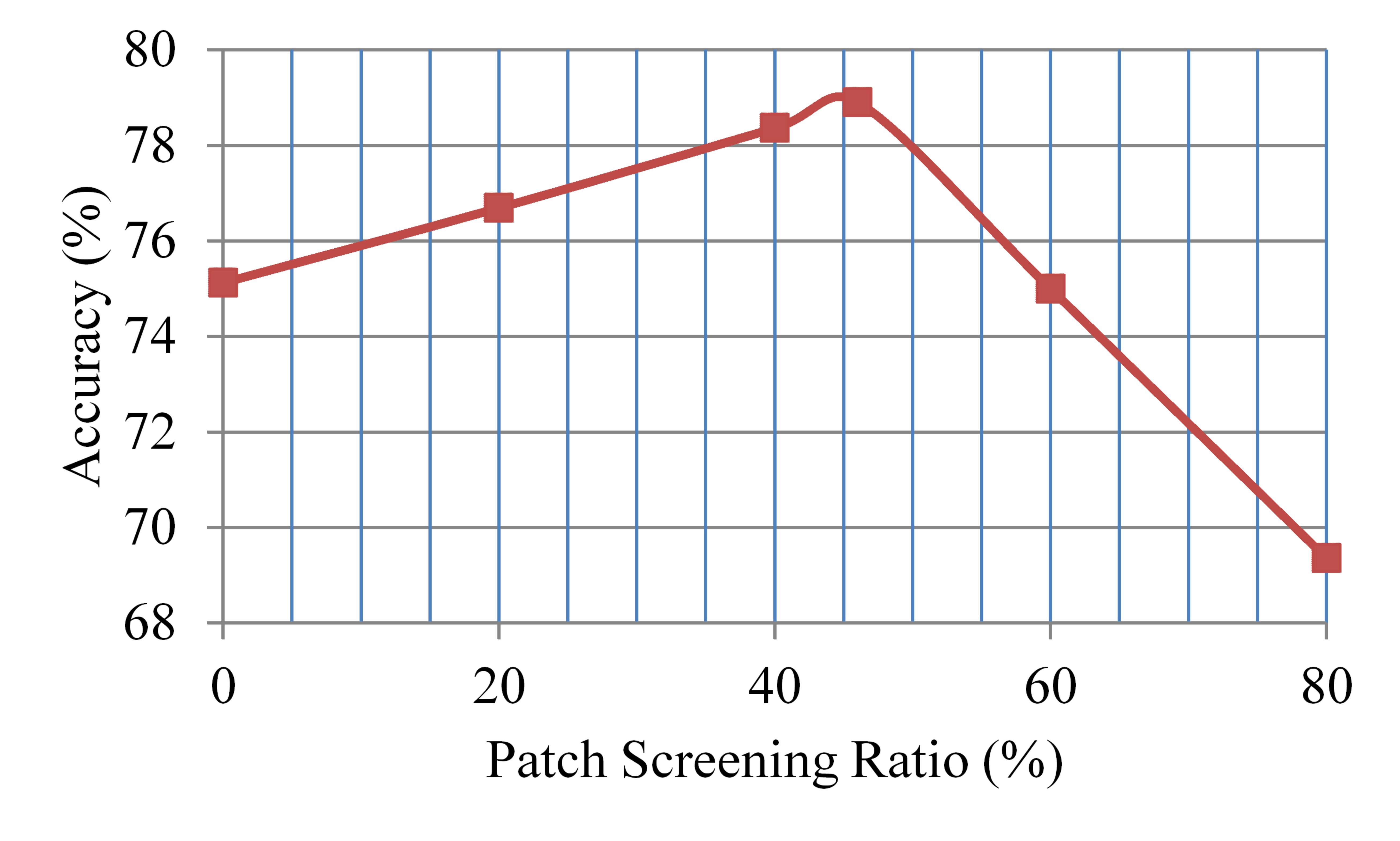}}
  \centerline{\includegraphics[width=3.0in]{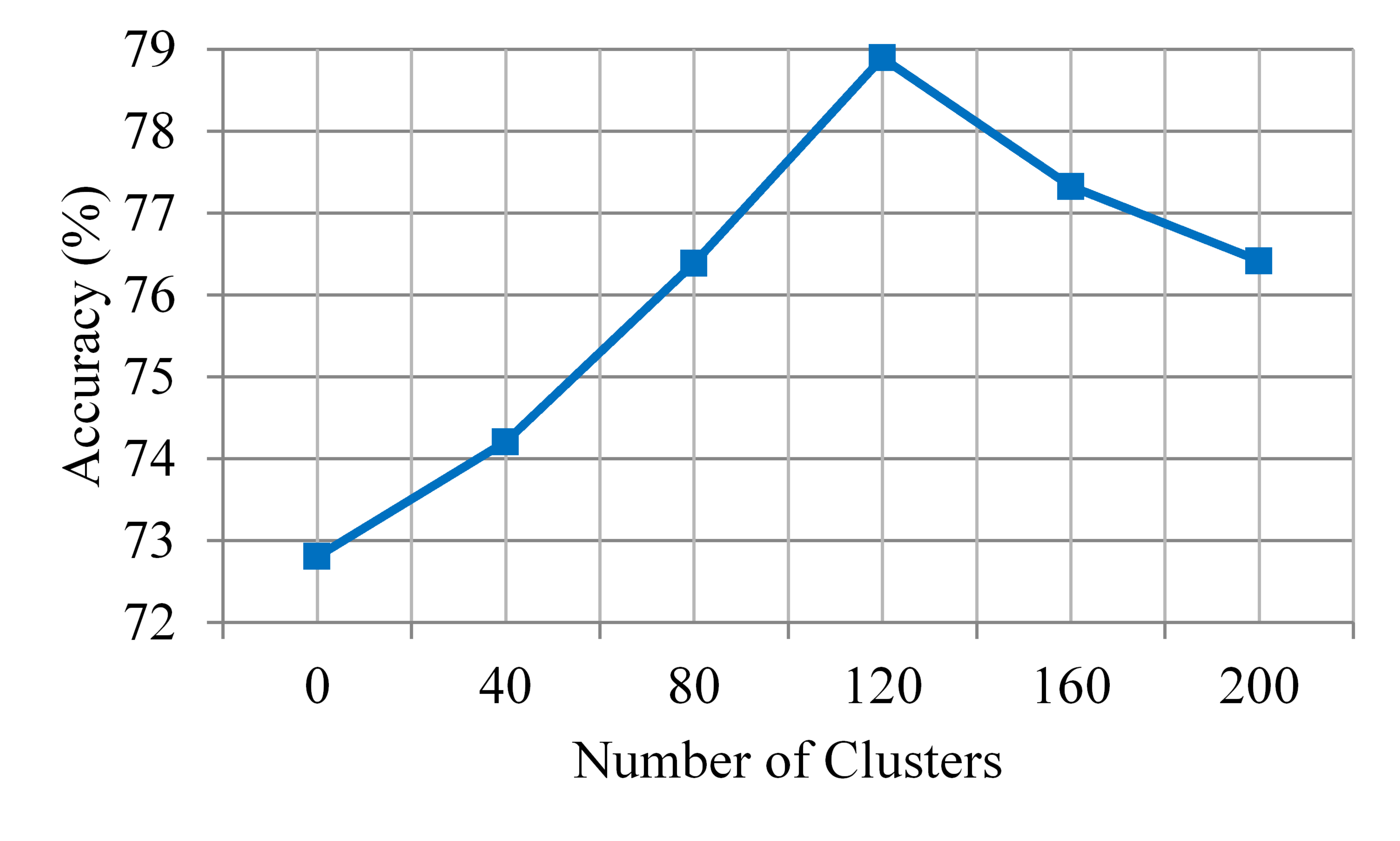}}
  \caption{\label{im:curve} Top: recognition accuracy vs. total screening ratio on MIT Indoor 67. Bottom: recognition accuracy vs. number of clusters in bottom level on MIT Indoor 67.}
\end{figure}

\paragraph{Importance of Outlier Removal and Patch Screening}
To see how important our outlier removal and patch screening stages are, one can directly feed all the object proposals without any screening into the subsequent components down the pipeline (discriminative clustering and local fine-tuning). During our patch screening step, as shown in Eq. \ref{eq.MIL}, we rank all the patches according to their discriminative weights and discard those with lower weights. Here we define the total screening ratio as the percentage of discarded patches in both outlier removal and patch soft screening steps. In Fig. \ref{im:curve} (top), we can see, when the total screening ratio is zero, the recognition accuracy is 75.12\% (also shown in the fifth row in Table. \ref{tab:inner}). This is because, although we have reasonable region proposals, there could still be many noisy ones among them. These noisy region proposals are either false positives or non-discriminative objects (as shown in Fig. \ref{im:flowgraph}) shared by multiple scene categories. On the other hand, an overly high screening ratio has also been found to hurt recognition performance, as shown in Fig. \ref{im:curve} (top). This is reasonably easy to understand because higher ratios could discard some discriminative meta objects that would otherwise contribute to the overall performance. We search for an optimal ratio through cross validation on a small subset of the training data. The outlier removal step is also important in filtering out regions that do not fit in a certain category and brings along 2.6\% improvement in final classification performance, as shown in the sixth row of Table. \ref{tab:inner}).

\vspace{-5mm}
\paragraph{Importance of Clustering} Next we justify the usefulness of clustering patches into meta objects. Without patch clustering, we can directly take the collection of screened patches as a large codebook, and treat every patch as a visual word. We then apply LSAQ \cite{Liu2011_LSAQ} (with 100 nearest neighbors) coding and SPM pooling to build the image-level representation. The resulting recognition rate on MIT Indoor 67 is 72.81\% (eighth row in Table. \ref{tab:inner}), which is around 6.1\% lower than the result of MetaObject-CNN. This controlled experiment demonstrates that patch clustering for meta object creation is crucial in our pipeline. Clustering patches into meta objects improves the generality and representativeness of those discovered discriminative patches because clustering emphasizes the common semantic meaning shared among similar patches while tolerating less important differences among them. Fig. \ref{im:curve} (bottom) shows the impact of the number of clusters in the bottom level on the final recognition rate. It is risky to group patches into an overly small number of clusters because it would assign patches with different semantic meanings to the same meta object. Creating too many clusters is also risky due to the poor generality of the semantic meanings of meta objects.

\paragraph{Importance of Local Fine-Tuning} Local fine-tuning has also proven to be effective in our pipeline. We tried using the responses from the RIM clustering model directly for pooling. On MIT Indoor 67, the recognition rate without local fine-tuning is 76.10\% (ninth row from the bottom in Table. \ref{tab:inner}), which is around 2.8\% lower than that with local fine-tuning. This demonstrates local fine-tuning actually defines better separation boundaries between clusters, which is consistent with the common sense about fine-tuning. We have also used the CNN locally fine-tuned on SUN397 to perform cross-dataset classification on MIT Indoor 67. The recognition rate is 76.52\% (bottom row in Table. \ref{tab:inner}), which indicates CNNs fine-tuned over one scene patch dataset have the potential to perform well on other scene datasets.

\section{Conclusions}
We have introduced a novel pipeline for scene classification, which is built on top of pre-trained CNN networks via explicitly harvesting discriminative meta objects in a local manner. Through extensive comparisons in a series of controlled experiments, our method generates state-of-the-art results on two popular yet challenging datasets,  MIT Indoor 67 and Sun397. Recent studies on convolutional neural networks, such as GoogLeNet \cite{Googlenet}, indicate that using deeper models would improve recognition performance more substantially than shallow ones. Therefore training better generic CNNs would certainly improve its transfer learning capability as well. Nevertheless, our approach is intrinsically orthogonal to this line of effort. Exploring other local fine-tuning methods would be an interesting direction for future work.

{\noindent \bf Acknowledgment}
This work was partially supported by Hong Kong Research Grants Council under General Research Funds (HKU17209714).

{\small
\bibliographystyle{ieee}
\bibliography{selectivePatch}

\begin{thebibliography}{10}\itemsep=-1pt

\bibitem{arandjelovic2013all}
R.~Arandjelovic and A.~Zisserman.
\newblock All about vlad.
\newblock In {\em CVPR}, pages 1578--1585. IEEE, 2013.

\bibitem{arbelaez2011contour}
P.~Arbelaez, M.~Maire, C.~Fowlkes, and J.~Malik.
\newblock Contour detection and hierarchical image segmentation.
\newblock {\em PAMI}, 33(5):898--916, 2011.

\bibitem{MCG:CVPR:14}
P.~A. Arbel{\'{a}}ez, J.~Pont{-}Tuset, J.~T. Barron, F.~Marqu{\'{e}}s, and
  J.~Malik.
\newblock Multiscale combinatorial grouping.
\newblock In {\em CVPR}, pages 328--335, 2014.

\bibitem{DBLP:conf/nips/DoerschGE13}
C.~Doersch, A.~Gupta, and A.~A. Efros.
\newblock Mid-level visual element discovery as discriminative mode seeking.
\newblock In {\em NIPS'13}, 2013.

\bibitem{Pedestrain:09}
P.~Dollar, C.~Wojek, B.~Schiele, and P.~Perona.
\newblock Pedestrian detection: An evaluation of the state of the art.
\newblock {\em TPAMI}, 34(4):743--761, April 2012.

\bibitem{Decaf}
J.~Donahue, Y.~Jia, O.~Vinyals, J.~Hoffman, N.~Zhang, E.~Tzeng, and T.~Darrell.
\newblock Decaf: {A} deep convolutional activation feature for generic visual
  recognition.
\newblock In {\em ICML}, pages 647--655, 2014.

\bibitem{RCNN:14}
R.~B. Girshick, J.~Donahue, T.~Darrell, and J.~Malik.
\newblock Rich feature hierarchies for accurate object detection and semantic
  segmentation.
\newblock In {\em CVPR}, 2014.

\bibitem{DBLP:conf/eccv/GongWGL14}
Y.~Gong, L.~Wang, R.~Guo, and S.~Lazebnik.
\newblock Multi-scale orderless pooling of deep convolutional activation
  features.
\newblock In {\em ECCV}, 2014.

\bibitem{Recognition:Region}
C.~Gu, J.~Lim, P.~Arbelaez, and J.~Malik.
\newblock Recognition using regions.
\newblock In {\em Computer Vision and Pattern Recognition, 2009. CVPR 2009.
  IEEE Conference on}, pages 1030--1037, June 2009.

\bibitem{jegou2010aggregating}
H.~J{\'e}gou, M.~Douze, C.~Schmid, and P.~P{\'e}rez.
\newblock Aggregating local descriptors into a compact image representation.
\newblock In {\em CVPR}, pages 3304--3311. IEEE, 2010.

\bibitem{juneja13blocks}
M.~Juneja, A.~Vedaldi, C.~V. Jawahar, and A.~Zisserman.
\newblock Blocks that shout: Distinctive parts for scene classification.
\newblock In {\em CVPR}, 2013.

\bibitem{DeepVideo14}
A.~Karpathy, G.~Toderici, S.~Shetty, T.~Leung, R.~Sukthankar, and L.~Fei{-}Fei.
\newblock Large-scale video classification with convolutional neural networks.
\newblock In {\em CVPR}, pages 1725--1732, 2014.

\bibitem{Krause2010}
A.~Krause, P.~Perona, and R.~G. Gomes.
\newblock Discriminative clustering by regularized information maximization.
\newblock In {\em NIPS}, 2010.

\bibitem{DBLP:conf/nips/KrizhevskySH12}
A.~Krizhevsky, I.~Sutskever, and G.~E. Hinton.
\newblock Imagenet classification with deep convolutional neural networks.
\newblock In {\em NIPS}, pages 1106--1114, 2012.

\bibitem{Lazebnik2006_SPM}
S.~Lazebnik, C.~Schmid, and J.~Ponce.
\newblock Beyond bags of features: Spatial pyramid matching for recognizing
  natural scene categories.
\newblock In {\em CVPR}, 2006.

\bibitem{ObjectBank}
L.~Li, H.~Su, Y.~Lim, and F.~Li.
\newblock Object bank: An object-level image representation for high-level
  visual recognition.
\newblock {\em IJCV}, 107(1):20--39, 2014.

\bibitem{Liu2011_LSAQ}
L.~Liu, L.~Wang, and X.~Liu.
\newblock In defense of soft-assignment coding.
\newblock In {\em ICCV}, 2011.

\bibitem{margolin2014otc}
R.~Margolin, L.~Zelnik-Manor, and A.~Tal.
\newblock Otc: A novel local descriptor for scene classification.
\newblock In {\em ECCV 2014}, pages 377--391. Springer, 2014.

\bibitem{DBLP:conf/cvpr/OquabBLS14}
M.~Oquab, L.~Bottou, I.~Laptev, and J.~Sivic.
\newblock Learning and transferring mid-level image representations using
  convolutional neural networks.
\newblock In {\em CVPR}, 2014.

\bibitem{DBLP:conf/iccv/PandeyL11}
M.~Pandey and S.~Lazebnik.
\newblock Scene recognition and weakly supervised object localization with
  deformable part-based models.
\newblock In {\em ICCV}, 2011.

\bibitem{Pfister14a}
T.~Pfister, K.~Simonyan, J.~Charles, and A.~Zisserman.
\newblock Deep convolutional neural networks for efficient pose estimation in
  gesture videos.
\newblock In {\em ACCV}, 2014.

\bibitem{MITIndoor67}
A.~Quattoni and A.~Torralba.
\newblock Recognizing indoor scenes.
\newblock In {\em CVPR}, pages 413--420, 2009.

\bibitem{DBLP:journals/corr/RazavianASC14}
A.~S. Razavian, H.~Azizpour, J.~Sullivan, and S.~Carlsson.
\newblock {CNN} features off-the-shelf: an astounding baseline for recognition.
\newblock {\em CoRR}, 2014.

\bibitem{sanchez2013image}
J.~S{\'a}nchez, F.~Perronnin, T.~Mensink, and J.~Verbeek.
\newblock Image classification with the fisher vector: Theory and practice.
\newblock {\em IJCV}, 105(3):222--245, 2013.

\bibitem{OCSVM:2001}
B.~Sch\"{o}lkopf, J.~C. Platt, J.~C. Shawe-Taylor, A.~J. Smola, and R.~C.
  Williamson.
\newblock Estimating the support of a high-dimensional distribution.
\newblock {\em Neural Comput.}, 13(7):1443--1471, July 2001.

\bibitem{DBLP:journals/corr/SermanetEZMFL13}
P.~Sermanet, D.~Eigen, X.~Zhang, M.~Mathieu, R.~Fergus, and Y.~LeCun.
\newblock Overfeat: Integrated recognition, localization and detection using
  convolutional networks.
\newblock {\em CoRR}, 2013.

\bibitem{Singh:2012:UDM:2403006.2403013}
S.~Singh, A.~Gupta, and A.~A. Efros.
\newblock Unsupervised discovery of mid-level discriminative patches.
\newblock In {\em ECCV 2012}, pages 73--86, 2012.

\bibitem{Googlenet}
C.~Szegedy, W.~Liu, Y.~Jia, P.~Sermanet, S.~Reed, D.~Anguelov, D.~Erhan,
  V.~Vanhoucke, and A.~Rabinovich.
\newblock Going deeper with convolutions.
\newblock {\em CoRR}, 2014.

\bibitem{SelectiveSearch:IJCV:13}
J.~R.~R. Uijlings, K.~E.~A. van~de Sande, T.~Gevers, and A.~W.~M. Smeulders.
\newblock Selective search for object recognition.
\newblock {\em International Journal of Computer Vision}, 104(2):154--171,
  2013.

\bibitem{icml2013_wang13d}
X.~Wang, B.~Wang, X.~Bai, W.~Liu, and Z.~Tu.
\newblock Max-margin multiple-instance dictionary learning.
\newblock In {\em ICML-13}, volume~28, May 2013.

\bibitem{Sun397}
J.~Xiao, J.~Hays, K.~Ehinger, A.~Oliva, and A.~Torralba.
\newblock Sun database: Large-scale scene recognition from abbey to zoo.
\newblock In {\em CVPR'10}, June 2010.

\bibitem{yang2009linear}
J.~Yang, K.~Yu, Y.~Gong, and T.~Huang.
\newblock Linear spatial pyramid matching using sparse coding for image
  classification.
\newblock In {\em CVPR}, pages 1794--1801. IEEE, 2009.

\bibitem{GigaSUN}
B.~Zhou, J.~Xiao, A.~Lapedriza, A.~Torralba, and A.~Oliva.
\newblock Learning deep features for scene recognition using places database.
\newblock In {\em NIPS}, 2014.

\bibitem{EdgeBox:ECCV:14}
C.~L. Zitnick and P.~Doll{\'{a}}r.
\newblock Edge boxes: Locating object proposals from edges.
\newblock In {\em ECCV}, 2014.

\end{thebibliography}
}

\end{document}